\documentclass[conference]{IEEEtran}
\IEEEoverridecommandlockouts

\usepackage{amsmath,amssymb,amsfonts}
\usepackage{algorithmic}
\usepackage{algorithmic,algorithm}
\usepackage{graphicx}
\usepackage{textcomp}
\usepackage{xcolor}
\usepackage{hyperref}
\usepackage{amsmath}
\usepackage{subfig}
\usepackage{biblatex}
\hypersetup{
  colorlinks   = true, 
  urlcolor     = red, 
  linkcolor    = blue, 
  citecolor    = blue 
}

\def\BibTeX{{\rm B\kern-.05em{\sc i\kern-.025em b}\kern-.08em
    T\kern-.1667em\lower.7ex\hbox{E}\kern-.125emX}}
    
\begin{document}

\title{ Vision-Based Intelligent Robot Grasping Using Sparse Neural Network \\}

\author{ 
\IEEEauthorblockN{ Priya Shukla\textsuperscript{1},  Vandana Kushwaha, G C Nandi }
\IEEEauthorblockA{ \textit{Center of Intelligent Robotics}\\
Indian Institute of Information Technology Allahabad,
Prayagraj-211015, U.P., India}
\IEEEauthorblockA{  priyashuklalko@gmail.com, kush.vandu@gmail.com, gcnandi@iiita.ac.in  }

}

\maketitle
\begingroup\renewcommand\thefootnote{*}
\footnotetext{Equal contributions from all the authors}
\endgroup
\begingroup\renewcommand\thefootnote{1}
\footnotetext{Corresponding Author}
\endgroup

\begin{abstract}
In the modern era of Deep Learning, network parameters plays a vital role in models efficiency but it has its own limitations like extensive computations and memory requirements, which may not be suitable for real time intelligent robot grasping tasks. Current research focuses on how the model efficiency can be maintained by introducing sparsity but without compromising accuracy of the model in robot grasping domain. More specifically, in this research two light-weighted neural networks have been introduced, namely Sparse-GRConvNet and Sparse-GINNet, which leverage sparsity in robotic grasping domain for grasp pose generation by integrating
the Edge-PopUp algorithm. This algorithm facilitates the identification of the top K\% of edges by considering their respective score values.
Both the Sparse-GRConvNet and Sparse-GINNet models are designed to generate high-quality grasp poses in real-time at every pixel location, enabling robots to effectively manipulate unfamiliar objects. We extensively trained our models using two benchmark datasets: Cornell Grasping Dataset (CGD) and Jacquard Grasping Dataset (JGD).
Both Sparse-GRConvNet and Sparse-GINNet models outperform the current state-of-the-art methods in terms of performance, achieving an impressive accuracy of 97.75\% with only 10\% of the weight of GR-ConvNet and 50\% of the weight of GI-NNet, respectively, on CGD. 
Additionally, Sparse-GRConvNet achieve an accuracy of 85.77\% with 30\% of the weight of GR-ConvNet and Sparse-GINNet achieve an accuracy of 81.11\% with 10\% of the weight of GI-NNet on JGD. To validate the performance of our proposed models, we conducted extensive experiments using the Anukul (Baxter) hardware cobot.


\end{abstract}


\section{\textbf{Introduction}}



Robotic grasping is essential for effective interactions between robots and the physical world. It involves the ability of robots to grasp objects accurately in dynamic and unstructured environments. Real-time grasping, with applications in industry, household robotics, and healthcare, has gained attention. However, intelligent grasping is complex, akin to human development where we learn skilled manipulation through hand-eye coordination. Recent advancements in machine learning, computer vision, and deep learning offer potential to create intelligent robot graspers.

These developments could lead to autonomous robots effectively interacting with the world. Challenges remain in creating efficient computational learning architectures \cite{hinton2022forward} 
with minimal trainable parameters \cite{ramanujan2020s,frankle2018lottery}.
Human brain learning \cite{neuroplasticity}, 
driven by neuro-plasticity, reshaping neural connections through experience, provides insights. Unlike fixed artificial neural networks, the brain adapts continuously, consuming minimal energy. While back-propagation learning persists, structure's role is emphasized, supported by rigorous experiments in predicting grasping rectangles for robots.

\begin{figure}[!ht]
\centering
\includegraphics[scale=0.35]{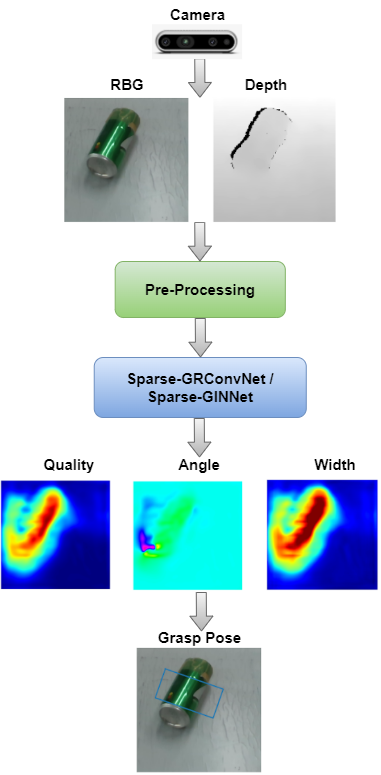}
\caption{ Overview of Grasp Pose Generation Process}
\label{overview}
\end{figure}

In this study, we have introduced a novel approach that incorporates the Edge-PopUp algorithm \cite{ramanujan2020s} into grasp generation models such as GR-ConvNet \cite{kumra2020antipodal} and GI-NNet \cite{shukla2022generative}. This integration has resulted in our proposed models, namely Sparse-GRConvNet and Sparse-GINNet, which effectively address the challenge of real-time grasping of unfamiliar objects.
Our proposed Sparse-GRConvNet and Sparse-GINNet models are characterized by their lightweight nature, as they have a significantly reduced number of parameters. Despite their lightweight design, these models achieve accuracy levels comparable to state-of-the-art models, making them suitable for real-time applications.
Overall, our work presents a practical solution that combines sparsity-based techniques, reduced parameterization, and high accuracy, enabling efficient real-time grasping of novel items.
Fig. \ref{overview}, shows the overview of grasp pose generation though the proposed models.

\begin{figure*}[!ht]
\centering
\includegraphics[scale=0.45]{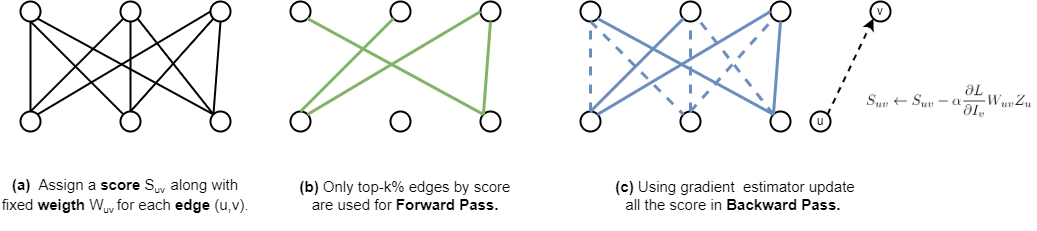}
\caption{ 
Edge-PopUp Algorithm \cite{ramanujan2020s}}
\label{edge}
\end{figure*}

In the Edge-PopUp algorithm, as illustrated in Fig. \ref{edge}, each edge in the network is assigned a positive real value known as a score, in addition to its weight. These scores play a crucial role in the selection of a subnetwork. Specifically, we identify the top K\% of edges with the highest scores and utilize their associated weights to construct the subnetwork.
During the backward pass, a gradient estimator 
is employed to update the scores of all edges. This dynamic update process enables previously inactive or ``dead" edges to become active again and rejoin the subnetwork. Importantly, only the score corresponding to each weight is updated during this process, while all other weights in the network remain unchanged.

We have evaluated the performances of our models  on both the benchmark dataset: CGD \cite{lenz2015deep} and, JGD \cite{depierre2018jacquard}. 
Both Sparse-GRConvNet and Sparse-GINNet
models outperform the current state-of-the-art methods in terms
of performance, achieving an impressive accuracy of 97.75\%
with only 10\% of the weight of GR-ConvNet and 50\% of the
weight of GI-NNet, respectively, on CGD.
Additionally, Sparse-GRConvNet achieve an accuracy of 85.77\%
with 30\% of the weight of GR-ConvNet and Sparse-GINNet
achieve an accuracy of 81.11\% with 10\% of the weight of GI-NNet on the JGD.
To validate the performance of our proposed models, we conducted extensive experiments using the Anukul (Baxter) hardware cobot with our test object set.

Here, is the summary of our paper's significant contributions:
\begin{itemize}
    \item[-] We have introduced the integration of sparsity into the grasp rectangle generation networks,  GR-ConvNet and GI-NNet, and named it Sparse-GRConvNet and Sparse-GINNet respectively.
    To the best of our knowledge, this is the first time we have leveraged the concept of sparsity  for the grasp pose prediction.

    \item[-] The Edge-PopUp algorithm is employed to choose the significant edges of GR-ConvNet and GI-NNet based on their assigned scores, leading to the creation of a more streamlined network known as Sparse-GRConvNet and Sparse-GINNet.
 
    \item[-] Our Sparse-GRConvNet and Sparse-GINNet models demonstrates a significant reduction in parameters. Despite their lightweight architecture, they achieve impressive levels of accuracy. 
    
    \item[-] To evaluate the effectiveness of our proposed models, we conducted comprehensive experiments utilizing the Anukul (Baxter) hardware cobot in conjunction with our designated test object set. 
\end{itemize}

\section{\textbf{Related Wor}k}

There is a lot of research being done to address the robot grasping issue. Finding a universal answer to this issue, however, is difficult since it calls for study in a variety of fields, and the existing limits of the hardware for grasping make it problematic to evaluate the suggested solutions. The two main kinds of grasping methods are described by Bohg et al. \cite{bohg2013data}  in more detail. The first is analytical methods, which demand accurate modeling of the gripper and the item. For instance, \cite{rodriguez2012caging} applies the idea of multi-finger grippers to the two-finger caging notion to ensure item entrapment. When the initial guess for the grasping locations and the 3-D object model is supplied, Krug et al. \cite{krug2010efficient}  describe an effective approach for determining the contact areas. 

The goal of data-driven techniques, in contrast, is to develop the ability to suggest effective grasping configurations for any type of item by using either external user input \cite{herzog2014learning}  or annotated training data \cite{jiang2011efficient}. The best-grasping candidate from the suggested choices is chosen using some heuristics or metrics. According to  \cite{herzog2014learning}, the best-matched object with a known grasp configuration is used to synthesize the grasping configurations for unknown things. By using kinesthetic training, the robot is taught how to grasp familiar things. Utilizing the knowledge from earlier grasp attempts, the grasp choice gets better with time. In \cite{jiang2011efficient}, Jiang et al. suggest representing grasping rectangles on the camera image plane to help detect suitable grasping stances. They provide a two-step approach based on this model to automatically locate the ideal orientated grasping rectangle for every given item.

Deep learning-based approaches have recently been developed to address the grasping problem \cite{caldera2018review, du2021vision}. Some of these methods make use of deep learning to forecast the optimal grasping stance based on sensory information. Annotated images are used by Lenz et al. \cite{lenz2015deep} to identify the ideal grasping stance. Their method involves rotating rectangles to identify the appropriate and inappropriate grasping stances for each object in the image. To serve as a comparison point for related strategies, this data is made available as the CGD. With the use of this data, they train a two-stage deep neural network that generates the grasping rectangle with the highest score. The network's initial stage develops the ability to suggest qualified grasping candidates. In contrast, the second step picks out the best-grasping rectangle by learning how to improve the candidates that were acquired. Similar to this, Redmon et al. \cite{redmon2015real} use the same dataset and suggest three distinct deep neural network designs to find candidates that are strong graspers. Their results outperform the strategy described in \cite{lenz2015deep}, which they used.

However, they don't offer any outcomes using a real robot. Using ResNet-50, a recent CNN network, as their foundation, Kumra et al. \cite{kumra2017robotic} describe the most current results on the CGD.
On top of ResNet-50, they have suggested several shallow topologies for unimodal (RGB only) and multimodal (RGBD) data. Their findings demonstrate that, on this benchmark dataset, the suggested unimodal and multimodal designs produce state-of-the-art outcomes. A deep neural network design for recognizing grasping rectangles from photos is also proposed by Guo et al. \cite{guo2016deep}. However, their method can only suggest rectangles that are horizontally aligned, which restricts its applicability to rotating objects. An approach that forecasts the grasp candidates' robustness is put forth by Mahler et al. \cite{mahler2017dex}. Three parameters: the grasp center coordinates, the vertical gripper angle, and the approach are used to establish a grasp position. They build a deep neural network to assess the robustness of a grasp candidate from depth photos using millions of artificial point clouds. Their approach makes use of parallel grippers and presupposes knowledge of the gripper's 3-D model. They tested their method on a real robot and had over 90\% success in grasping.

In \cite{morrison2018closing, morrison2020learning}, Morrison et al.  propose GG-CNN and GG-CNN2, a fully convolutional neural network that predicts the antipodal grasp pose for each pixel. Building upon this, Kumra et al. \cite{kumra2020antipodal} further enhance the model's capabilities by incorporating residual modules and named their model as GR-ConvNet. 
Although GG-CNN has a significantly smaller number of parameters compared to GR-ConvNet, it is important to note that GR-ConvNet outperforms GG-CNN in terms of accuracy for both benchmark datasets.
 Priya et al. in \cite{shukla2022generative} proposed a lightweight model called GI-NNet, which is based on GG-CNN. They incorporated an inception module into the architecture of the model. The GI-NNet model achieved higher accuracy compared to GR-ConvNet on the CGD while also reducing the number of parameters required.
 In other recent work \cite{mahajan2020robotic, shukla2021development}, the researchers proposed a novel discriminative-generative model that combines the representation quality of VQ-VAE with GG-CNN and GG-CNN2, named RGGCNN and RGGCNN2 respectively. This integrated model has improved the grasp pose prediction, especially in scenarios with limited labeled data. By leveraging the strengths of these different components, the model offers enhanced performance and robustness in grasp pose estimation. 
 \cite{kushwaha2023generating} To address the challenge of the limited availability of labeled grasping datasets, a generative-based model has been proposed. This model aims to generate grasp poses for both seen and unseen objects. 

Unlike previous approaches, our research introduces a groundbreaking concept of sparsity to grasp pose prediction models such as GR-ConvNet and GI-NNet.
This novel approach allows us to develop a lightweight network that achieves comparable or even superior accuracy compared to the current state-of-the-art methods.

\section{\textbf{Problem Formulation}}

In this study, the problem of robotic grasping, is defined as the prediction of  grasp for objects in a given scene.

The grasp pose in the robot's frame is denoted by (\ref{g_r}):
\begin{equation}
    G_r= ( P, \theta_r, W_r, Q)  
    \label{g_r}
\end{equation}
where,   $P=(x, y, z)$ refers to end-effector's center position,  $\theta_r$ represents its rotation  around the z-axis,  $W_r$ represents the required width for it, and $Q$ denotes the  quality score for grasp.

The predicted grasp pose   for an n-channel image $I=R^{n\times h \times w}$ with a height of $h$ and width of $w$ is denoted as (\ref{g_i}):

\begin{equation}
    G_i= ( x, y, \theta_i, W_i, Q)  
    \label{g_i}
\end{equation}
Here, $(x, y)$: the grasp's center and $W_i$: the required width  are in the image's frame, $\theta_i$ represents the rotation in the camera's frame of reference,  and $Q$ remains the same scalar as mentioned above.

To transform the predicted grasp pose ($G_i$), which is in the image coordinate plane, into the robot coordinate frame ($G_r$), we apply a series of transformations using (\ref{transform}). 

\begin{equation}
    G_r= T^r_c \Big( T^c_i ( G_i) \Big)  
    \label{transform}
\end{equation}
 where a grasp pose $G_i$ in image space is transformed using the $T^c_i$    into the camera's 3D space, and then   the camera space is transformed into the robot space using $T^r_c$. 

Our aim is to create a lightweight network that has the capability to predict the optimal grasp pose for unfamiliar objects based on an n-channel image of the scene.

\begin{figure*}[!ht]
\centering
\includegraphics[scale=0.35]{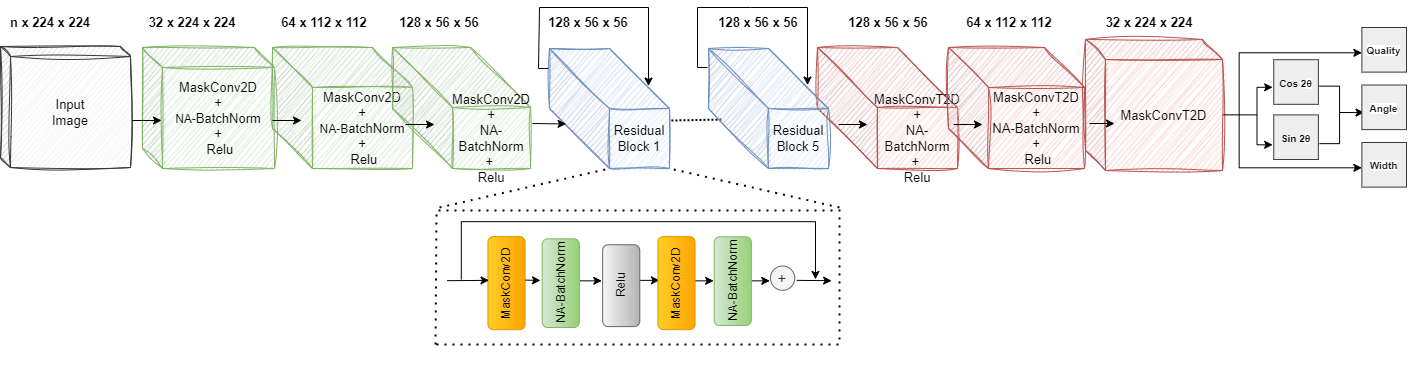}
\caption{ 
Architecture of Sparse-GRConvNet}
\label{sgrconv}
\end{figure*}

\section{\textbf{ Methodology}}

In this part, we explain our methodology and start with a quick summary.

\subsection{\textbf{Preliminaries}}

 In our study, we have adopted GR-ConvNet as the base architecture for Sparse-GRConvNet and GI-NNet as the base architecture for Sparse-GINNet. Furthermore, we have integrated the Edge-PopUp algorithm into our framework to enhance the grasp generation process. In this section, we will provide a comprehensive description of the architecture of GR-ConvNet, GI-NNet, and the Edge-PopUp algorithm.

\subsubsection{\textbf{GR-ConvNet and GI-NNet}}

GR-ConvNet and GI-NNet are generative grasping networks specifically designed to generate grasp poses for each pixel in an input image captured by a camera. These networks are designed to accommodate input images with dimensions of $n\times224\times224$. Notably, the input modality is not limited to a specific type, such as depth-only or RGB-only images. Instead, GR-ConvNet and GI-NNet can effectively process input images with an arbitrary number of channels, making them suitable for various input modalities. This flexibility enables the models to be versatile and capable of handling different types of input data.

The GR-ConvNet architecture  consists of three convolutional layers, five residual layers, and transposed convolution layers.
The convolutional layers are responsible for extracting relevant features from the input image, allowing the network to identify important patterns. The residual layers further enhance the network's ability to capture fine details and comprehend complex patterns in the input image. Finally, the transposed convolution layers upsample the features and generate the final output images. 
 However, the GI-NNet architecture comprise of three 2D convolutional layers for feature extraction, followed by five inception blocks that select different filter sizes in parallel to reduce computation. Transposed convolution layers are then used for upsampling, and a final convolution operation generates the desired output images. This architecture efficiently captures contextual information, reduces complexity, and produces accurate grasp pose predictions.

The output of both GR-ConvNet and GI-NNet consists of four images, each representing specific aspects of the grasp pose for every pixel in the input image. These images correspond to grasp quality, angle (represented as sin$2\theta$ and cos$2\theta$), and grasp width. Each pixel in these output images contains information about the grasp quality, angle, and width at that particular location in the input image.
By utilizing the generated output images, both networks can predict the grasp pose for each pixel, providing a detailed representation of the grasp pose for the object present within the input image. These predicted grasp poses offer valuable information for robotic manipulation and can be used to guide robots in effectively grasping objects.

 However, the large number of parameters in GR-ConvNet (19,00,900) makes it less suitable for real-time applications. While GI-NNet offers a reduced parameter count of 5,92,300, it is still relatively higher compared to other \cite{morrison2018closing, morrison2020learning}, which have significantly fewer parameters (approximately 62K and 66K, respectively).


\subsubsection{\textbf{Edge-PopUp Algorithm}}

In \cite{ramanujan2020s}, researchers have presented a concept suggesting the presence of potential subnetworks within an over-parameterized neural network that achieves comparable performance. They have developed an algorithm named as Edge-PopUp for selecting the relevant edge for creating a subnetwork.

\begin{algorithm}[!htbp]
    \caption{  Edge-PopUp Algorithm}
    \textbf{Input:}
    Image I, Score $S_{uv}$ and Weights $W_{uv}$ for the  Edge $(u,v)$ of the Network \\
    \textbf{Output:} After training a potential subnetwork is obtained with edges of  top\%K score.\\
    \textbf{Procedure:}
    \label{edge-algo}
    \begin{algorithmic}[1]
        \STATE First  initialized the network parameters like scores, weights, and biases for each layer.
        \STATE Use the edge corresponding to the top K\%  scores for the forward pass in  each layer and  compute the gradient.
        \STATE In backward pass update all the scores with the gradient estimator.
        \begin{equation}
           S_{uv} \leftarrow S_{uv} - \alpha \frac{\partial L}{\partial I_v} {W_{uv}Z_u}
        \end{equation}
        where, ${W_{uv}Z_u}$ denotes the weighted output of neuron $u$, and $I_v$ denotes the input of neuron $v$.
        \STATE Repeat till accuracy is updated.
\end{algorithmic}
\end{algorithm}
In Algorithm \ref{edge-algo}, steps for  the Edge-PopUp algorithm are described. Initially, the neural network parameters, specifically the weights, are initialized along with an additional parameter called ``score" for each edge in every layer. During the forward pass, only the top K\% of edges are selected based on their score values. Subsequently, during the backward pass, all score values are updated using the gradient estimator.
Through this dynamic update process, previously inactive or ``dead" edges have the opportunity to become active once again and rejoin the subnetwork. It is important to note that only the score associated with each weight is updated, while all other weights in the network remain unchanged.

After the training process, a potential subnetwork, also known as a sparse network, is obtained that demonstrates a comparable performance to the original fully connected network.

\begin{figure*}[!ht]
\centering
\includegraphics[scale=0.35]{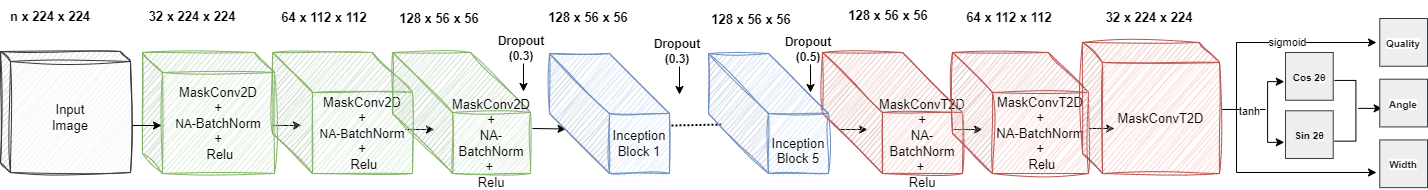}
\caption{ 
Architecture of Sparse-GINNet }
\label{sginet}
\end{figure*}

\begin{figure}[!ht]
\centering
\includegraphics[scale=0.5]{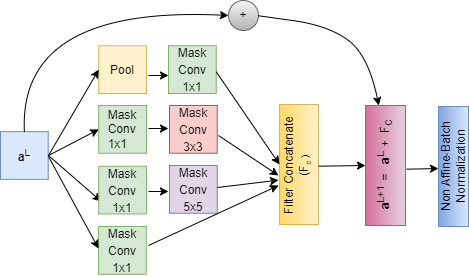}
\caption{ 
Inception block of Sparse-GINNet }
\label{inception}
\end{figure}

\subsection{\textbf{Proposed Approach}}
Here, we have explored the integration of the Edge-PopUp algorithm with GR-ConvNet and GI-NNet to develop our proposed models, Sparse-GRConvNet and Sparse-GINNet.
This integration allows us to leverage the benefits of sparsity, resulting in lightweight networks with reduced computational requirements. 

\subsubsection{\textbf{Architecture of Sparse-GRConvNet and Sparse-GINNet}}
  For visual representations of the proposed models, namely Sparse-GRConvNet and Sparse-GINNet, refer to  Fig. \ref{sgrconv} and Fig. \ref{sginet} respectively. The key modification involves replacing the  convolutional layers with mask-convolutional layers and substituting transposed convolution layers with transposed mask-convolutional layers within their respective base architectures. Furthermore, the original batch normalization is replaced with non-affine batch normalization. The mask-convolutional layers and transposed mask-convolutional layers serve as wrapper classes for the corresponding convolutional layers, incorporating the Edge-PopUp algorithm. This algorithm is responsible for selecting the most relevant edges based on their top score values, enhancing the sparsity and computational efficiency of both the models. Fig. \ref{inception} shows the inception block of Sparse-GINNet.

Both Sparse-GRConvNet and Sparse-GINNet models takes an input image of size $224 \times 224$ with n channels and produces three output images: a quality image,  angle images (for sin$2\theta$ and cos$2\theta$), and a width image, all representing pixel-wise grasp information. By utilizing these output images, the model predicts the grasp pose for the object.
The overall parameter count of the Sparse-GR-ConvNet and Sparse-GINNet varies depending on the chosen sparsity value (K). Different sparsity values lead to changes in the number of parameters throughout the model.

\subsubsection{\textbf{Training Details}}
During the training process, RGB-D images are utilized, and various train-test splits of the dataset are employed for different sparsity values. The training is conducted with a batch size of 8, using the Adam optimizer with a learning rate of 0.001. The Sparse-GRConvNet model run for a total of 50 epochs, while the Sparse-GINNet model run for 30 epochs.

\section{\textbf{Performance Evaluation}}

\subsection{\textbf{Datasets}}
For both the training and testing of our models, we utilized two benchmark datasets:  Cornell Grasping Dataset and Jacquard Grasping Dataset. 

\subsubsection{\textbf{Cornell Grasping Dataset (CGD)}}

The Cornell Grasping Dataset \cite{lenz2015deep} is a comprehensive dataset that contains RGB-D images of various real objects. It consists of 885 RGB-D photos, capturing 240 different objects. The dataset provides annotations for positive and negative grasps, with a total of 5,110 positive grasps and 2,909 negative grasps. The annotations are represented as rectangular bounding boxes with pixel-level coordinates, depicting antipodal grasps.
To augment the dataset, we employed random cropping, zooming, and rotation techniques, resulting in approximately 51,000 grasps. During training, we only considered positively labeled grasps from the dataset.
 
\subsubsection{\textbf{Jacquard Grasping Dataset (JGD)}}

The JGD \cite{depierre2018jacquard} is created using a portion of ShapeNet, a substantial dataset of CAD models. This dataset focuses on effective gripping positions and includes annotations generated from grasp attempts conducted in a simulated environment.
It consists of 54,000 RGB-D photos, each accompanied by annotations indicating the locations of effective gripping positions on the objects. These annotations were derived from the grasp attempts performed within the simulated setting, resulting in a total of 1.1 million occurrences of successful grasps. With such a significant number of grip occurrences and a large collection of RGB-D photos, this dataset offers comprehensive coverage of grasping scenarios and object variations, and no need for data augmentation during model training. 

\subsection{\textbf{Grasp Detection Metric}}


To ensure a fair comparison of our model's performance, we have used the rectangle metric \cite{jiang2011efficient}  proposed by Jiang et al. 
As per the rectangle metric, a grasp is considered good if it satisfies the following two criteria:

\begin{itemize}
    \item The intersection over union (IoU) score between the predicted  and the ground truth grasp rectangle should be higher than 25\%.
    \item The orientation offset between the ground truth and predicted grasping rectangles should be less than $30^{\circ}$. 
\end{itemize}

\subsection{\textbf{Implementation}}

\subsubsection{\textbf{Setup}}
\begin{figure}[!ht]
\centering
\includegraphics[scale=0.12]{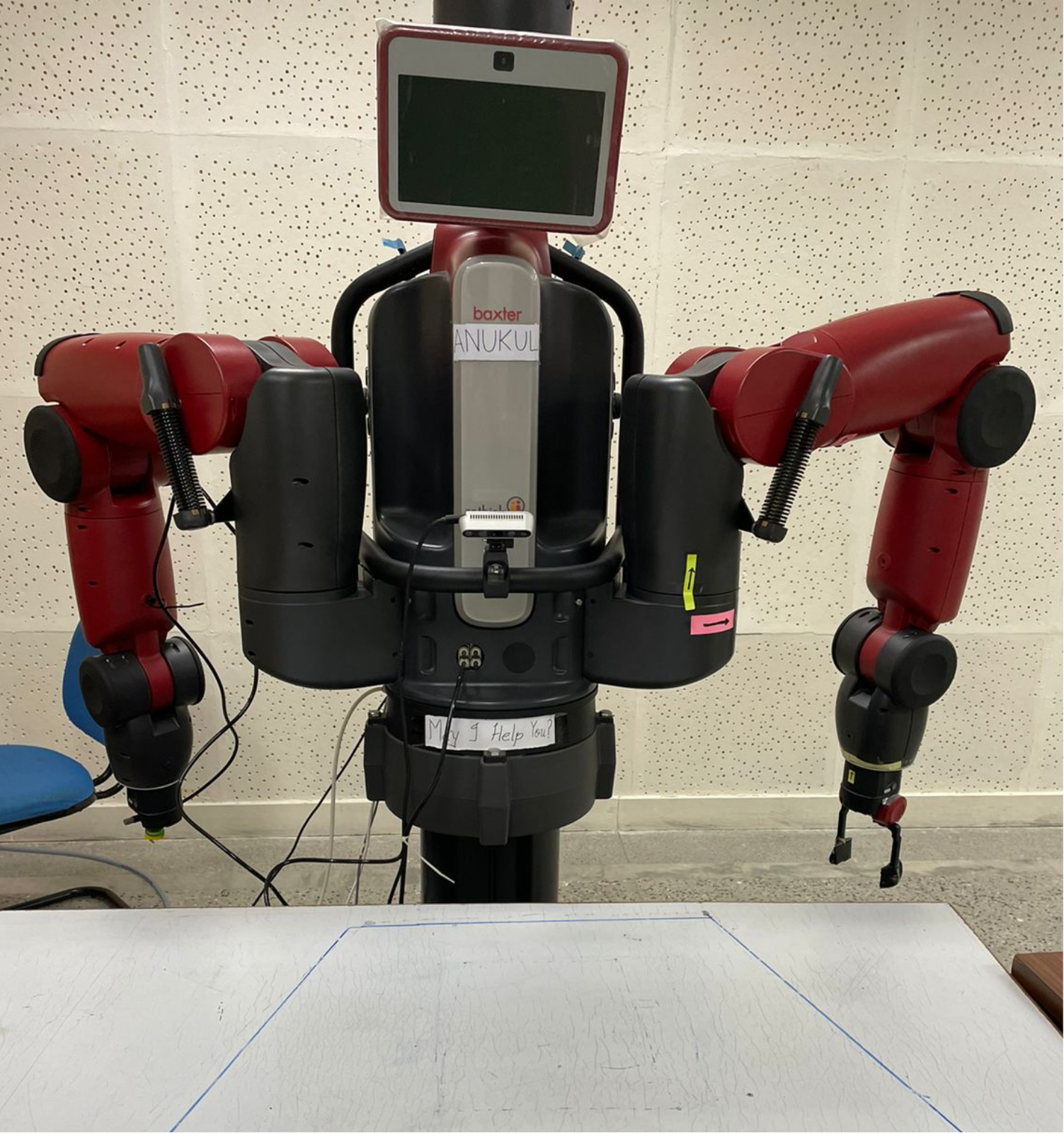}
\caption{Experimental Setup.}
\label{baxter}
\end{figure}

Our trial setup consists of Anukul, a Baxter Cobot build by Rethink Robotics, equipped with two arms that have seven degrees of freedom. Additionally, an externally attached high-resolution stereo camera (Intel RealSense D435) is positioned at the torso of Anukul, as depicted in Fig. \ref{baxter}. For our real-time grasping experiments, we have employed a parallel plate gripper with two fingers, allowing Anukul to grasp objects effectively.

\subsubsection{\textbf{Procedure to perform robotic grasping}}
First, the RGB-D data captured by the external camera serves as input to the trained model of Sparse-GRConvNet and Sparse-GINNet, that predicts the grasp pose for the object in the scene. The highest-scoring grasp pose is selected as the target grasping point for execution. 

Then,  the selected grasp pose ($G_i$), which is in the image coordinate plane, is transform to the robot coordinate frame using (\ref{transform}) to obtain ($G_r$). 
After computing $G_r$ in the robot coordinate frame, joint angle for Anukul's arm is computed using  an inverse kinematic solution  and passed to the control system of Anukul to execute the grasping.

\begin{figure}[!ht]
\centering
\includegraphics[scale=0.28]{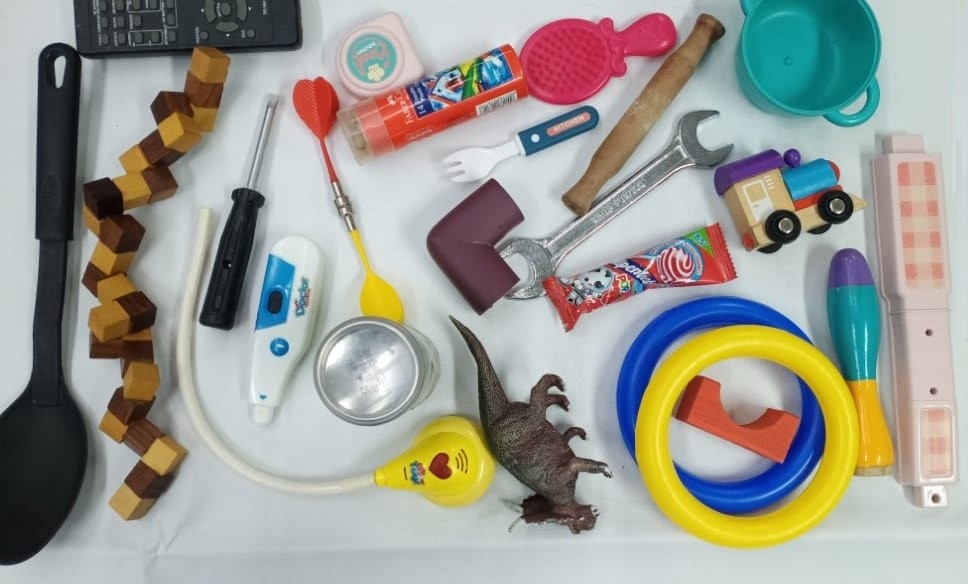}
\caption{Test objects used during experiments.}
\label{test_object}
\end{figure}

\subsection{\textbf{Results}}
In this section, we have discussed the results of our experiments. We assessed the effectiveness of the Sparse-GRConvNet and Sparse-GINNet on two benchmark datasets: CGD and JGD. Additionally, we evaluated our model's performance on a test object set, as depicted in Fig. \ref{test_object}.

\subsubsection{\textbf{Cornell Grasping Dataset (CGD)}}

We have conducted a performance evaluation of  Sparse-GRConvNet and  Sparse-GINNet  on the CGD, considering various sparsity (K) values  (10\%, 30\%, 50\%, 70\%, 90\%) for different dataset split ratios (10-90, 30-70, 50-50, 70-30, 90-10). TABLE \ref{gr_cornell} summarizes the accuracy achieved by  the Sparse-GRConvNet model across different train/test split ratios   and sparsity  values. The accuracy achieved by  the Sparse-GINNet model for different train/test split ratios and sparsity values is tabulated in TABLE \ref{gi_cornell}.

\begin{table}[!ht]
\caption{Sparse-GRConvNet's accuracy on the CGD with varying sparsity values and train-test split ratios.}
\centering
\begin{tabular}{|c|c|c|c |c|c|}
\hline
\textbf{$\downarrow$ Sparsity Value }& \multicolumn{5}{|c|}{\textbf{Accuracy (\%) on  different split ratio of Dataset}} \\[3pt]
\hline
\textbf{(K\% of Weight) }& {(10-90)} & {(30-70)} & {(50-50)} &{(70-30)} & {(90-10)} 
\\[3pt]
\hline 
$10 $ & $81.30$ & $87.25$ & $93.67$ & $92.85$ & $\textbf{97.75}$ \\[3pt]
\hline
$30 $ & $80.05$ & $88.22$ & $92.55$ & $91.72$ & $\textbf{97.75}$         \\[3pt]
\hline
$50 $ & $79.54$ & $91.61$ & $88.03$ & $92.10$ & $\textbf{96.62}$   
\\[3pt]                                        
\hline
$70$ &  $72.39$ & $87.25$ & $91.87$ & $86.46$ & $\textbf{92.13}$       \\[3pt]
\hline 
$90 $ & $62.35$ & $49.51$ & $51.46$ & $56.01$ & $\textbf{62.92} $        \\[3pt]
\hline
\end{tabular}
\label{gr_cornell}
\end{table}

\begin{table}[!ht]
\caption{Sparse-GINNet's accuracy on the CGD with varying sparsity values and train-test split ratios.}
\begin{center}
\begin{tabular}{|c|c|c|c |c|c|}
\hline
\textbf{$\downarrow$ Sparsity Value }& \multicolumn{5}{|c|}{\textbf{Accuracy (\%) on  different split ratio of Dataset}} \\[3pt]
\hline
\textbf{(K\% of Weight) }& {(10-90)} & {(30-70)} & {(50-50)} &{(70-30)} & {(90-10)} 
\\[3pt]
\hline 
$10 $   & $87.57$ & ${91.77}$ & ${90.97}$         & ${91.72}$        & $\textbf{95.50}$        \\[3pt]
\hline
$30 $   &  $76.78$         & $90.16$          & ${93.22}$  & $90.97$          & $\textbf{96.62}$          \\[3pt]
\hline
$50 $   &  $75.78$         & ${87.58}$        & $92.32$          & $87.21$          & $\textbf{97.75}$ \\[3pt]
\hline
$70$    &  $85.19$         & $83.54$          & $86.68$          & $\textbf{93.60}$ & $92.13$          \\[3pt]
\hline 
$90 $   &  $43.28$         & $44.67$          &  $44.69$         &  $56.39$         & $\textbf{62.92} $        \\[3pt]
\hline
\end{tabular}
\label{gi_cornell}
\end{center}
\end{table}

\begin{figure*}[!ht]
    \centering
    \subfloat[][Sparse-GRConvNet with 10\% split]
    {\includegraphics[scale=0.31]{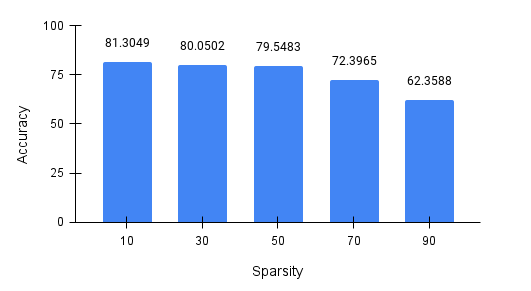}
    \label{a}}
    \hspace{0cm}
    \subfloat[][Sparse-GRConvNet with 30\% split]
    {\includegraphics[scale=0.31]{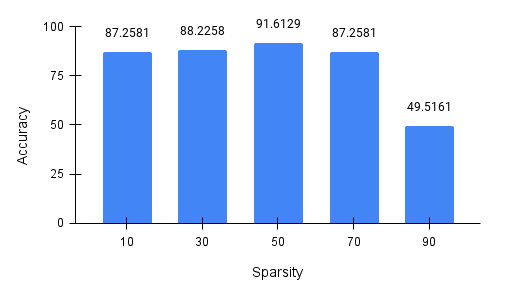}
    \label{b}}
    \hspace{0cm}
    \subfloat[][Sparse-GRConvNet with 50\% split]
    {\includegraphics[scale=0.31]{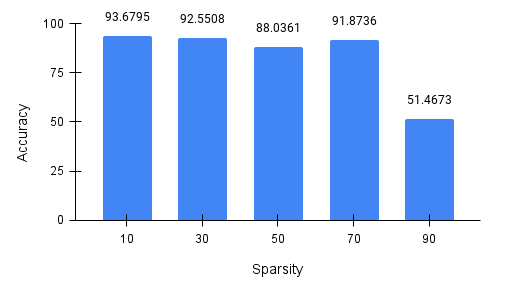}
    \label{c}}
    \hspace{0cm}
    \subfloat[][Sparse-GRConvNet with 70\% split]
    {\includegraphics[scale=0.31]{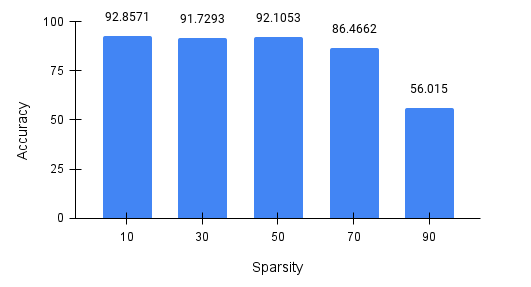}
    \label{d}}
    \hspace{0cm}
    \subfloat[][Sparse-GRConvNet with 90\% split]
    {\includegraphics[scale=0.31]{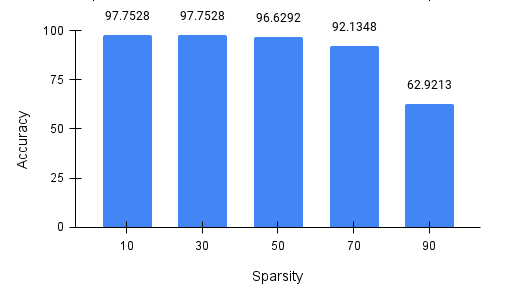}
    \label{e}}
    \caption{Performance of Sparse-GRConvNet on CGD for different sparsity values on different split ratios.
    }
    \label{sgrcon_cornell_histo}
\end{figure*}

\begin{figure*}[!ht]
    \centering
    \subfloat[][Sparse-GINNet with 10\% split]
    {\includegraphics[scale=0.31]{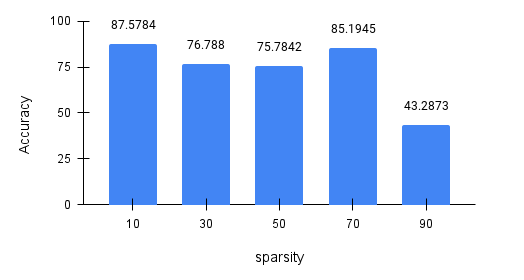}
    \label{aa}}
    \hspace{0cm}
    \subfloat[][Sparse-GINNet with 30\% split]
    {\includegraphics[scale=0.31]{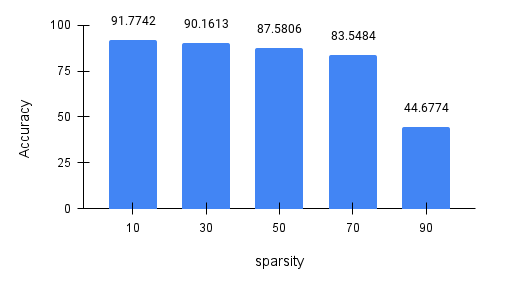}
    \label{bb}}
    \hspace{0cm}
    \subfloat[][Sparse-GINNet with 50\% split]
    {\includegraphics[scale=0.31]{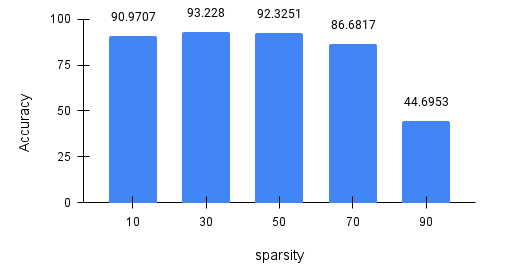}
    \label{cc}}
    \hspace{0cm}
    \subfloat[][Sparse-GINNet with 70\% split]
    {\includegraphics[scale=0.31]{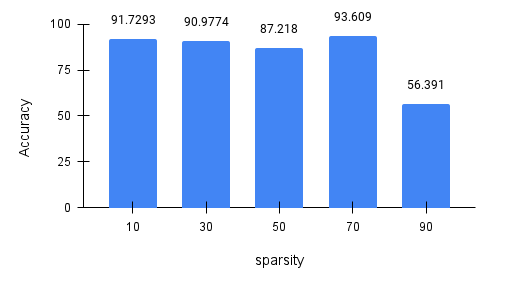}
    \label{dd}}
    \hspace{0cm}
    \subfloat[][Sparse-GINNet with 90\% split]
    {\includegraphics[scale=0.31]{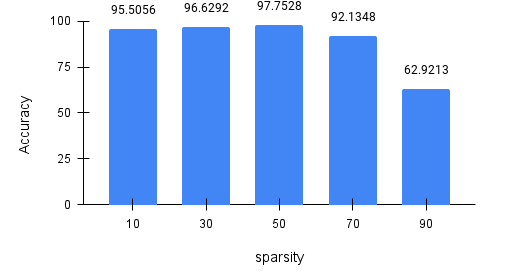}
    \label{ee}}
    \caption{Performance of Sparse-GINNet on CGD for different sparsity values on different split ratios.
    }
    \label{sginnet_cornell_histo}
\end{figure*}

\begin{table}[!ht]
\caption{Comparison of our proposed models on CGD at different sparsity values.}
\begin{center}
\scalebox{0.86}{
\begin{tabular}{|c|c|c|c|c|}
\hline
 \textbf{Model $\rightarrow$} &  \multicolumn{2}{|c|}{\textbf{Sparse-GRConvNet}} &  \multicolumn{2}{|c|}{ \textbf{Sparse-GINNet}} \\[3pt]
\hline
\textbf{Sparsity (K) $\downarrow$} &  \textbf{Parameters}  & \textbf{Accuracy (\%)} & \textbf{Parameters}  & \textbf{Accuracy (\%)} \\[3pt]
\hline
K=10\%   & $1,90,090$    & $\textbf{97.75}$  & $59, 230$     & ${95.50}$ \\[3pt]
\hline
K=30\%   & $5,70,270$    & $\textbf{97.75}$  & $1,77,690$   & $96.62$ \\[3pt]
\hline
K=50\%   & $9,50,450$    & $96.62$           & $2,96,150$   & $\textbf{97.75}$ \\[3pt]
\hline
K=70\%   & $13,30,630$  & $92.13$            & $4,14,610$   &  ${93.60}$\\[3pt]
\hline
K=90\%   & $17,10,810$  & $62.92$            & $5,33,070$   &   $62.92 $  \\[3pt]
\hline
\end{tabular}}
\label{sgr_gi}
\end{center}
\end{table}

\begin{table}[!ht]
\caption{Comparative study of proposed models with their base models on CGD for varied train-test splits.}
\begin{center}
\scalebox{0.85}{
\begin{tabular}{|c|c|c|c|c|}
\hline
 \textbf{Train-Test} &  \textbf{GR-ConvNet} & \textbf{Sparse-GRConvNet} & \textbf{GI-NNet} & \textbf{Sparse-GINNet} \\[3pt]
 \hline
 \textbf{Split} & \textbf{(base) } & \textbf{(ours)} & \textbf{(base)} & \textbf{(ours)}\\[3pt]
\hline
10-90 & $87.64$ & $81.30$ & $89.88$ & $87.57$ \\[3pt]
\hline
30-70 & $89.88$ & $91.61$ & $95.50$ & $91.77$ \\[3pt]
\hline
50-50 & $96.62$ & $93.67$ & $96.62$ & $93.22$ \\[3pt]
\hline
70-30 & $95.50$ & $92.85$ & $98.87$ & $93.60$\\[3pt]
\hline
90-10 & $97.75$ & $97.75$ & $98.87$ & $97.75$ \\[3pt]
\hline
\end{tabular}}
\label{base1}
\end{center}
\end{table}

\begin{table}[!ht]
\caption{ Proposed models comparison with their base models on CGD.}
\begin{center}
\scalebox{0.99}{
\begin{tabular}{|c|c|c|}
\hline
\textbf{Model }         &  \textbf{Parameters} & \textbf{Accuracy (\%)} \\[3pt]
\hline
GR-ConvNet  (base)      & $19,00,900$           & $97.75$  \\[3pt]
\hline
Sparse-GRConvNet (ours)  & $1,90,090$           & $97.75$  \\[3pt]
\hline
GI-NNet (base)           & $5,92,300$            &  $98.87$ \\[3pt]                      
\hline
Sparse-GINNet (ours)     & $2,96,150$            & $97.75$ \\[3pt]
\hline
\end{tabular}}
\label{base2}
\end{center}
\end{table}

\begin{figure}[!ht]
\centering
\includegraphics[scale=0.53]{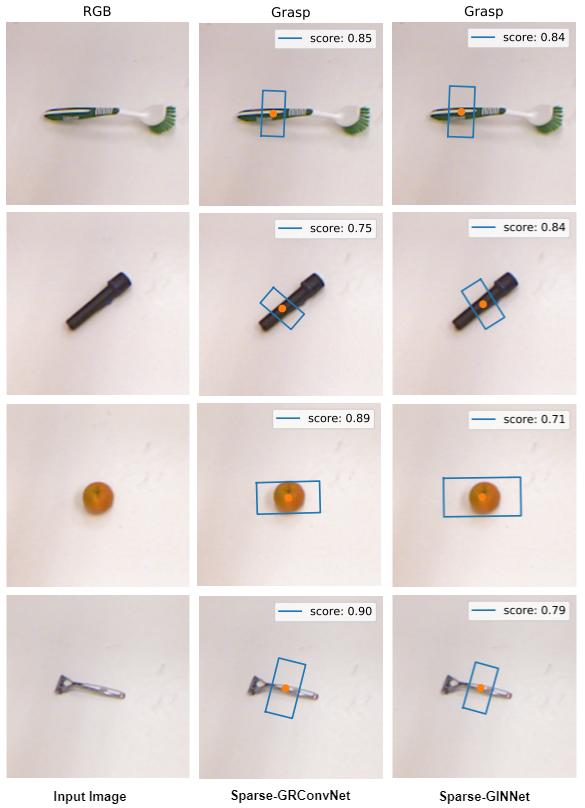}
\caption{Performance visualisation by both models for input images from the CGD.}
\label{cornell_result}
\end{figure}

\begin{table}[!ht]
\caption{Performance of trained models with CGD.}
\begin{center}
\scalebox{1.0}{
\begin{tabular}{|c|c|c|}
\hline
\textbf{Authors} & \textbf{Model}& \textbf{ Accuracy (\%)} \\ [2pt]
& & (Image-Wise)\\ [3pt]
\hline
Jiang et al.\cite{jiang2011efficient} & Fast Search & $60.5$  \\[3pt]
\hline
Morrison et al.\cite{morrison2018closing} &  GGCNN  & $73.0$ \\[3pt]
\hline
Lenz et al.\cite{lenz2015deep} & SAE, struct. reg. & $73.9$ \\[3pt]
\hline
Wang et al.\cite{zeng2022robotic} & Two-stage closed-loop &  $85.3$ \\[3pt]
\hline 
Redmon et al.\cite{redmon2015real} & AlexNet, MultiGrasp & $88.0$ \\[3pt]
\hline
Asif et al.\cite{asif2018ensemblenet} & STEM-CaRFs & $88.2$ \\[3pt]
\hline
Karaoguz et al.\cite{karaoguz2019object} & GRPN & $88.7$ \\[3pt]
\hline
Kumra et al.\cite{kumra2017robotic} & ResNet-50x2 & $89.2$ \\[3pt]
\hline
Asif et al.\cite{asif2018graspnet} & GraspNet & $90.2$ \\[3pt]
\hline
Guo et al.\cite{guo2016deep} & ZF-net &  $ 93.2$ \\[3pt]
\hline
Zhou  et al.\cite{zhou2018fully}& FCGN, ResNet-101 & $ 97.7$ \\[3pt]
\hline
Kumra et al.\cite{kumra2020antipodal} & GR-ConvNet & $97.7$ \\[3pt]
\hline
Shukla  et al.\cite{shukla2022generative} & GI-NNet  & $98.8$ \\[3pt]
\hline
Ours & Sparse-GRConvNet & $97.7$ \\[3pt]

 & Sparse-GINNet & $97.7$ \\[3pt]
\hline
\end{tabular}}
\label{finall}
\end{center}
\end{table}

Fig. \ref{sgrcon_cornell_histo} and Fig. \ref{sginnet_cornell_histo} display the histogram representation corresponding to TABLE \ref{gr_cornell} and TABLE \ref{gi_cornell}, respectively. From both Figures \ref{sgrcon_cornell_histo} and \ref{sginnet_cornell_histo},  we can conclude that using full weights might not be a  good idea. We are getting better results with less number of weight in most of the cases. 
TABLE \ref{sgr_gi} shows the comparison of both proposed model's parameters and the best accuracy achieved by them for different sparsity on the CGD. While, TABLE \ref{base1} shows the overall  performances of the proposed models in comparison to their base model, which shows that our models have achieved remarkable accuracy with the reduced number of network parameters.

As per TABLE \ref{base2}, Sparse-GRConvNet has attained an accuracy of  \textbf{$97.75$\%} that is equivalent to its base model but with reduced  parameters, count of $1,90,090$  which is only  $10$\% weight of the  base model. Whereas,  Sparse-GINNet has achieved an accuracy of \textbf{$97.75$\%} that is comparable to its base model that too with less number of parameters $2,96,150$ that is only $50$\% weight of the base model. From the obtained result it is evident that our models are lightweight and appropriate to be applied in real-time application.
In TABLE \ref{finall}, we have compared the performance of our proposed models with existing state-of-the art methods on the CGD. Fig. \ref{cornell_result} shows the performance visualization of both proposed models for some input images from the CGD.

\begin{figure}[!ht]
\centering
\includegraphics[scale=0.53]{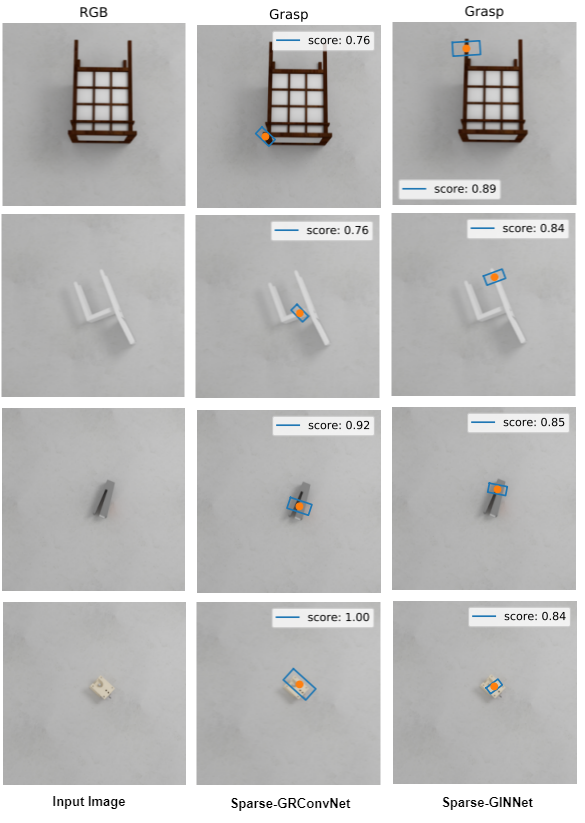}
\caption{Performance visualisation by both models  for input images from the JGD.}
\label{jacq_result}
\end{figure}

\begin{figure*}[!ht]
\centering
\includegraphics[scale=0.6]{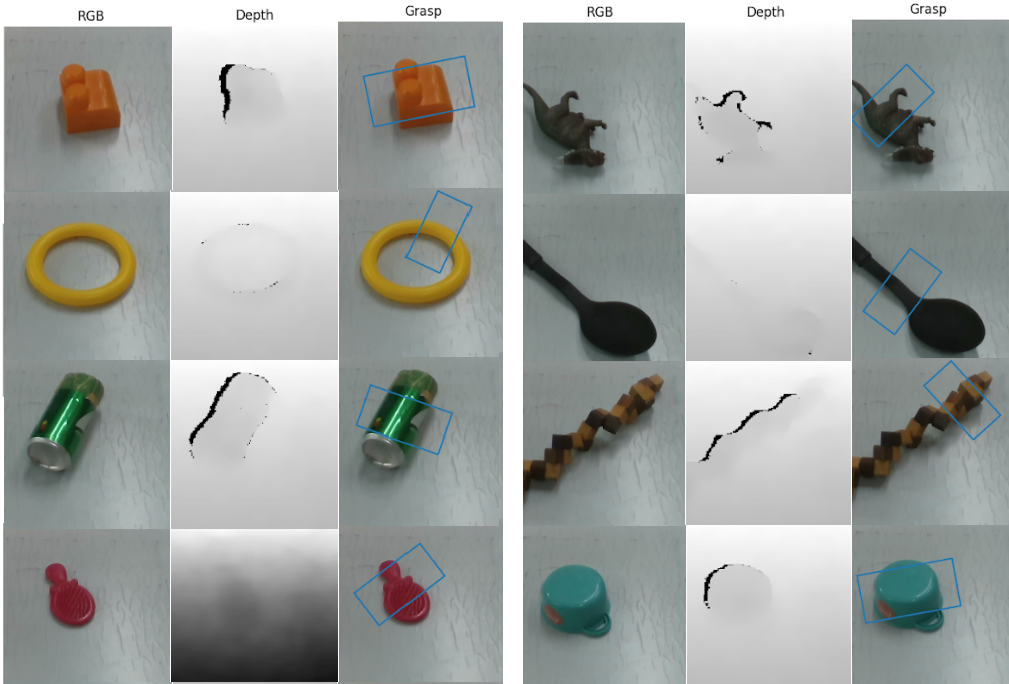}
\caption{Performance visualisation with proposed models on test objects.}
\label{object_set}
\end{figure*}

\subsubsection{\textbf{Jacquard Grasping Dataset (JGD)}}

\begin{table}[!ht]
\caption{Comparison of our proposed models on JGD with various sparsity values.}
\begin{center}
\scalebox{0.88}{
\begin{tabular}{|c|c|c|c|c|}
\hline
\textbf{Model $\rightarrow$} &  \multicolumn{2}{|c|}{\textbf{Sparse-GRConvNet}} &  \multicolumn{2}{|c|}{ \textbf{Sparse-GINNet}}\\[3pt]
\hline
\textbf{Sparsity (K) $\downarrow$} &  \textbf{Parameters}  & \textbf{Accuracy (\%)} & \textbf{Parameters}  & \textbf{Accuracy (\%)}  \\[3pt]
\hline
K=$10$\% & $1, 90, 090$ & $73.84$  & $59, 230$ & $\textbf{81.11}$\\[3pt]
\hline
K=$30$\% & $5, 70, 270$ & $\textbf{85.77}$ & $1, 77, 690$  & $78.89$\\[3pt]
\hline
K=$50$\% & $9, 50, 450$ & $82.98$ & $2, 96, 150$ & $76.01$ \\[3pt]
\hline
K=$70$\% & $13, 30, 630$ & $77.57$ & $4, 14, 610$ & $71.16$  \\[3pt]
\hline 
K=$90$\% & $17, 10, 810$ & $44.54$ & $5, 33, 070$  & $39.31$ \\[3pt]
\hline
\end{tabular}}
\label{jacq}
\end{center}
\end{table}

\begin{table}[!ht]
\caption{Comparison of models' performance on JGD.}
\begin{center}
\scalebox{0.89}{
\begin{tabular}{|c|c|c|c|}
\hline
\textbf{Author} &  \textbf{Model } & \textbf{Accuracy (\%)} & \textbf{Parameters}  \\[3pt]
\hline
Morrison et al.\cite{morrison2020learning} & GG-CNN2              & $84$     & $66,000$           \\[3pt]
\hline
Kumra et al. \cite{kumra2020antipodal} & GR-ConvNet              & $94.6$     & $19,00,900$           \\[3pt]
\hline
Ours & Sparse-GRConvNet  & $85.75$   & $5,70,270$             \\[3pt]

& Sparse-GINNet          & $81.11$    & $59, 230$            \\[3pt]
\hline
\end{tabular}}
\label{final_j}
\end{center}
\end{table}

TABLE \ref{jacq} shows the performances of the proposed models on  JGD. The Sparse-GRConvNet and Sparse-GINNet models achieved accuracies of 85.77\% and 81.11\%, respectively,
on JGD with split ratio (90-10). These results were obtained by utilizing 30\% of the weights from the GR-ConvNet model for Sparse-GRConvNet and 10\% of the weights from the GI-NNet model for Sparse-GINNet. 
In TABLE \ref{final_j}, we can observe a performance comparison in terms of achieved accuracy and model parameters between the proposed model and other existing methods. Fig. \ref{jacq_result} shows the performance visualization of both proposed models on input images from the JGD.

\subsubsection{\textbf{Results on Test object set}}
Fig. \ref{object_set}  shows the performance    visualization of both proposed models on the test object set.

\section{Conclusion}
Designing a slim yet highly accurate deep neural network architecture is crucial, especially for real-time applications like robot grasping. 
We argued that searching for an appropriate neural network structure is equally important as deriving an ad hoc architecture based solely on empirical insights and attempting to determine weight parameters using the back propagation algorithm.  We have introduced the Sparse-GRConvNet and Sparse-GINNet
models by integrating an edge-selecting algorithm with grasp generation models like GR-ConvNet
and GI-NNet to harness the benefits of sparsity. These models exhibits a remarkable reduction in parameters while maintaining comparable accuracy. We evaluated the performance of our model on two benchmark datasets, namely CGD and JGD, and achieved significant accuracy. To explore the impact of sparsity, we trained our model using different data split ratios. 
Both Sparse-GRConvNet and Sparse-GINNet
models outperform the current state-of-the-art methods in terms
of performance, achieving an impressive accuracy of 97.75\%
with only 10\% of the weight of GR-ConvNet and 50\% of the
weight of GI-NNet, respectively, on CGD.
Additionally, Sparse-GRConvNet achieve an accuracy of 85.77\%
with 30\% of the weight of GR-ConvNet and Sparse-GINNet
achieve an accuracy of 81.11\% with 10\% of the weight of GI-NNet on the JGD which is a remarkable improvement.
Future endeavors should aim to either eliminate or minimize the reliance on the backpropagation algorithm. This could involve replacing it with suitable structures and conducting parameter tuning through the application of associative rules, such as those rooted in Hebb's law.   



\section*{Acknowledgements}
The present research is partially funded by the I-Hub foundation for Cobotics (Technology Innovation Hub of IIT-Delhi set up by the Department of Science and Technology, Govt. of India). Some of the experimental results presented here were undertaken by our undergraduate and dual degree students including Prasanth Kota, Mridul Mahajan and others during their semester long projects and otherwise. 


\printbibliography

\end{document}